\documentclass{Interspeech2024}

\interspeechcameraready 
\usepackage{devanagari}
\usepackage{tipa}
\usepackage{xcolor}
\usepackage{float}
\usepackage{soul}
\usepackage{booktabs}
\usepackage{graphicx}
\usepackage{multirow}

\title{Rasa: Building Expressive Speech Synthesis Systems for Indian Languages in Low-resource Settings}

\name[affiliation={}]{Praveen}{Srinivasa Varadhan$^*$}
\name[affiliation={}]{Ashwin}{Sankar$^*$}
\name[affiliation={}]{Giri}{Raju}
\name[affiliation={}]{Mitesh}{M. Khapra}

\address{AI4Bharat, Indian Institute of Technology Madras, India}
\email{cs21d201@cse.iitm.ac.in, \{ashwins1211, girirajur023\}@gmail.com, miteshk@cse.iitm.ac.in\\}


\begin{document}
\maketitle
\def\thefootnote{*}\footnotetext{Equal contribution.}\def\thefootnote{\arabic{footnote}}

\begin{abstract}
We release Rasa, the first multilingual expressive TTS dataset for any Indian language, which contains 10 hours of neutral speech and 1-3 hours of expressive speech for each of the 6 Ekman emotions covering 3 languages: Assamese, Bengali, \& Tamil. Our ablation studies reveal that just 1 hour of neutral and 30 minutes of expressive data can yield a Fair system as indicated by MUSHRA scores. Increasing neutral data to 10 hours, with minimal expressive data, significantly enhances expressiveness. This offers a practical recipe for resource-constrained languages, prioritizing easily obtainable neutral data alongside smaller amounts of expressive data. We show the importance of syllabically balanced data and pooling emotions to enhance expressiveness. We also highlight challenges in generating specific emotions, e.g., fear and surprise.
\end{abstract}

\section{Introduction}

Development of expressive TTS systems is a costly and time-consuming affair as it requires (i) the creation of recording scripts with emotion-laced content (ii) skilled voice artists proficient in conveying diverse emotional nuances (iii) adherence to studio-grade recording quality (iv) strict quality control measures to ensure fidelity of intended emotions and alignments between audio and transcripts. Given these requisites, it is not a surprise that there are only a few publicly available datasets for expressive TTS \cite{Busso2008IEMOCAP, lemoine2020Att, Zhou2021Seen, Cui2021EMOVIE, nguyen23, Cowen2017}, and the ones which are available cover only English, French, and Mandarin Chinese. Moreover, several of these datasets cover a restricted range of emotions, failing to cover even the six basic Ekman emotions \cite{ekman1992argument}. As a result, the academic community does not get enough opportunity to conduct research on expressive TTS synthesis. 
A case in point is that of Indian languages, wherein TTS datasets with \textit{neutral} voices are publicly available \cite{baby2016resources, Black2019CMU, Srivastava2020IndicSpeech}, but no expressive TTS datasets are available for any Indian language. 

To address this gap, we introduce, \textit{Rasa}, an expressive TTS dataset for Assamese, Bengali, and Tamil, which belong to two different language families.
It covers all 6 basic Ekman emotions with 1-3 hours of data per emotion and 10 hours of neutral data from each language. To build this dataset, we first created expressive recording scripts in each of these languages with the help of Large Language Models as well as human annotators. 
Using \textit{Rasa}, we build expressive TTS systems using FastPitch \cite{lancucki2021fastpitch}, an unsupervised alignment learning framework \cite{badlani2022one}, and HiFiGAN-V1 \cite{kong2020hifi}. 
We conduct detailed ablation studies for Tamil by varying the amount of neutral and expressive data, to understand the data requirements for building expressive TTS systems for Indian languages. 
We find that by using syllabically balanced data, we can build a Fair TTS system using just 1 hour of neutral data and 30 minutes of expressive data per emotion. The quality of the TTS system further improves as we add more neutral and expressive data per emotion. We replicate these studies for Assamese and Bengali, along with a small scale study for Hindi. 

In summary, the main contributions of our work are (i) a first-of-its-kind publicly available expressive TTS dataset for three Indian languages, (ii) expressive TTS systems built using this dataset, and (iii) detailed ablation studies leading to insights and a practical recipe for data collection for building expressive TTS systems for low-resource languages. We hope that
the insights from our work on the distribution of neutral and expressive data across emotions may help in making informed choices for data collection in other Indian languages. 
All the code, datasets, and models built in this work will be publicly released and enable further research on expressive TTS systems.

\section{Rasa: A High-Quality Expressive Speech Synthesis Dataset}
\label{sec:rasa}

Below we describe (i) the process for creating scripts for recording expressive content, (ii) the setup and process used for recording audio for these scripts with the help of a professional artist, and (iii) the quality control process used for ensuring that the resulting dataset is of high quality.

\subsection{Generating Recording Scripts}
A primary challenge in curating recording scripts for the task at hand is the lack of publicly available emotion-labeled text corpora in Indian languages. 
To overcome this challenge, we rely on the following two strategies. 

\subsubsection{LLM-Assisted Corpora Creation}
\label{sec:llm_assisted}
We prompt large language models (LLMs) like ChatGPT and WriteSonic to generate sentences in a single emotion across a variety of domains such as health, finance, sports, education, technology, etc., and encourage conversational outputs. We also add guidelines in our prompts to generate sentences that are simple to translate to Indian languages and oriented in first-person, because we often found this conveys emotion better when spoken. 
In this manner, we generate around 1800 English sentences (300 per emotion). 
We retained sentences that conveyed the intended emotion as certified by human annotators. The retained sentences were subsequently translated into the 3 languages by human translators, who were asked to prioritize fidelity of emotion over fidelity of translation, \textit{i.e.}, they had the flexibility of making semantic adjustments to improve expression, and omitting sentences with poor emotional intensity. 

\subsubsection{Scenario-prompted Human-Assisted Corpora Creation}
While LLM-assisted sentence generation proved effective, it needed careful prompting to ensure diversity and avoid repetition or mixed emotions. 
To enhance the corpus further, we set up a team of native writers who were asked to imagine expressive sentences in everyday Indian scenarios spanning various life situations. For instance, role-play scenarios like a conversation between a banker and a customer seeking loan provide opportunities to craft expressions of happiness or anger.

\subsubsection{Reusing Existing Corpora for Neutral Sentences}
In addition to expressive sentences, we also need neutral sentences to efficiently train TTS systems. For this, we selected utterances from 3 sources - (i) IndicTTS database \cite{baby2016resources}, (ii) Sangraha \cite{khan2024indicllmsuite}, and (iii) BPCC-H-Wiki \cite{gala2023}, which contain a diverse range of neutral-like utterances from multiple domains. 

\subsection{Recording Audio}
\textbf{Audio Recordings:} All audio files were recorded in a professional studio environment  using RØDE NT1 5th Generation Large-Diaphragm Studio Condenser Microphone, complete with a pop filter.
The audio recordings are stored in WAV format, maintaining specific standards: a sample rate of \SI{48}{\kilo\hertz}, a bit depth of 16 bits, and mono channel configuration. \\
\textbf{Voice Artist Selection:} 
We looked for native speakers who spoke with a local accent in the standard dialect of the language as used in government broadcasts. 
To find the most suitable voice, we asked the potential artists to send us 2-5 minutes of recordings where they expressed different emotions. These recordings were evaluated by native speakers, and the artist who won the majority preference from the listeners was chosen. 
\\
\textbf{Ethical Considerations:} The terms of data collection were reviewed and approved by our Institute Ethics Committee. 
The artists were clearly explained about the purpose of data collection, and were paid as per industry rates. Before the recording sessions, the voice artist signed a consent form, indicating their agreement to lend their voice and release the resulting data under permissible licenses for research and commercial purposes. They were also informed that the data will be used to train AI models which can synthesise new sentences in their voice. 
They felt no risk of losing their job as their work typically requires them to dub in multiple different character voices, as opposed to their standard voice used in our recordings.
\subsection{Quality Control}
We manually filter out profanity and toxicity from the recording scripts, and rectify typos and grammatical errors. 
Additionally, we reject 2075 sentences that didn't accurately convey the intended emotion. Similarly, on the audio front, we trimmed periods of silence before and after the utterance and manually normalized amplitude levels for consistent sound. Apart from this, we ensure the audio sample is not digitally altered in any way. To maintain naturalness, we re-recorded or eliminated recordings with irregular pauses, stuttering, and mispronunciations that are not aligned with the intended emotion. 

\begin{table}[!t]
\centering
\fontsize{8pt}{10pt}\selectfont
\caption{Data statistics of Rasa an expressive TTS dataset for Assamese (as), Bengali (bn), and Tamil (ta).}
\begingroup
\setlength{\tabcolsep}{2pt} 
\renewcommand{\arraystretch}{1} 
\begin{tabular}{@{}lcccrrrrrr@{}}
\toprule
\multirow{2}{*}{\textbf{Emotion}} & \multicolumn{3}{c}{\textbf{Durations (Hours)}}                                                      & \multicolumn{3}{c}{\textbf{\# Unique Syllables}}                                                    & \multicolumn{3}{c}{\textbf{\# Utterances}}                                                          \\ \cmidrule(l){2-4} \cmidrule(l){5-7} \cmidrule(l){8-10}
                                  & \multicolumn{1}{c}{\textbf{as}} & \multicolumn{1}{c}{\textbf{bn}} & \multicolumn{1}{c}{\textbf{ta}} & \multicolumn{1}{c}{\textbf{as}} & \multicolumn{1}{c}{\textbf{bn}} & \multicolumn{1}{c}{\textbf{ta}} & \multicolumn{1}{c}{\textbf{as}} & \multicolumn{1}{c}{\textbf{bn}} & \multicolumn{1}{c}{\textbf{ta}} \\ \midrule
Happy                             & 1.1                             & 1.3                             & 3.2                             & 631                             & 748                             & 998                             & 579                             & 669                             & 2342                            \\
Sad                               & 1.3                             & 1.3                             & 3.1                             & 579                             & 705                             & 883                             & 479                             & 635                             & 2283                            \\
Angry                             & 1.1                             & 1.0                             & 3.0                             & 578                             & 721                             & 920                             & 448                             & 603                             & 2387                            \\
Fear                              & 1.1                             & 1.1                             & 3.1                             & 590                             & 721                             & 867                             & 499                             & 614                             & 2226                            \\
Disgust                           & 1.1                             & 1.1                             & 3.2                             & 604                             & 730                             & 887                             & 489                             & 665                             & 2353                            \\
Surprise                          & 1.0                             & 1.2                             & 3.1                             & 590                             & 751                             & 957                             & 461                             & 637                             & 2311                            \\
Neutral                           & 10.4                            & 10.0                            & 9.5                             & 1761                            & 1944                            & 1595                            & 4251                            & 4788                            & 3559                            \\ \midrule
\textbf{Total}                    & 17.1                            & 17.0                            & 28.2                            & 5333                            & 6320                            & 7107                            & 7206                            & 8611                            & 17461                           \\ \bottomrule
\end{tabular}

\endgroup
\label{tab:rasa-dataset-statistics}
\end{table}

\subsection{Dataset Statistics}
A breakdown of the statistics of the expressive TTS dataset for Assamese, Bengali, and Tamil is presented in Table \ref{tab:rasa-dataset-statistics}. 
Syllable coverage, obtained using the Indic NLP Library \cite{kunchukuttan2020indicnlp}, reflects the linguistic coverage of the dataset.

\section{Methodology}
\subsection{Model Architecture}
We follow the design choices proposed in \cite{Kumar2022Towards} for building TTS systems for Indian languages. Specifically, we adopt FastPitch \cite{lancucki2021fastpitch}, a parallel text-to-speech model based on the Transformer architecture \cite{vaswani2017attention}. 
Instead of using external durations to train the duration predictor, we use an unsupervised alignment learning framework \cite{badlani2022one} to get alignments between the textual and acoustic features.  To support emotional conditioning, we add learnable emotion embeddings to the text encoder outputs. To synthesize the final audio waveforms from mel-spectrograms we use HiFiGAN-V1 \cite{kong2020hifi} trained from scratch. 

\subsection{Syllabic-Balanced Text Selection}
In low-resource settings, the careful selection of texts is pivotal.
At a much coarser level, ensuring sufficient word coverage would translate well to balanced phonotactic coverage in the selected corpora. However, achieving this in a low-resource setup is difficult due to the limited availability of unique words. We thus choose a middle ground and ensure diversity at the syllable level instead of the phoneme level or word level. We use the Indic NLP library \cite{kunchukuttan2020indicnlp} to obtain syllables in a given word.
We then use a variant of the greedy text selection algorithm \cite{santen1997methods}, that chooses the shortest candidate sentence containing the least common syllable at every greedy step. We empirically find that this variant of the algorithm helps well in ensuring sufficient coverage of less-frequent syllables while maintaining the relative frequency distribution of more frequent syllables.

\section{Experimental Setup}
\textbf{Dataset:} We use \textit{Rasa}\footnote{\url{https://github.com/AI4Bharat/Rasa}}, to conduct all our experiments. \\ 
\textbf{Implementation:}
Our models are implemented using the Coqui-TTS library and trained on a single NVIDIA A100 40GB GPU. We use the Adam optimizer with $\beta_1=0.99$ and $\beta_2=0.998$, a weight decay of $\lambda=10^{-6}$, and train the models 
until convergence. We use the Noam learning rate schedule \cite{vaswani2017attention}, initial learning rate of $0.0001$, and a batch size of 16. \\
\textbf{Evaluation:}
We assess the performance of our models using both objective and subjective metrics on a validation set and use 50 and 120 previously unseen utterances for neutral and expressive TTS, respectively. For acoustic evaluation, we employ two metrics: mel-cepstral distortion (MCD) \cite{kubichek1993mel} and the root-mean-square error of the logarithm of fundamental frequencies ($F_0$), while applying dynamic time warping \cite{salvador2007toward}. To gauge intelligibility, we employ the character error rate (CER), using text extracted from publicly available ASR models\footnote{\url{https://github.com/AI4Bharat/Indic-Conformer}}. To subjectively measure expressiveness of speech we employ MUSHRA \cite{series2014method} tests (Exp-MUSHRA) with 22 human raters using webMUSHRA \cite{schoeffler2018webmushra} and report the average $\pm$ 95\% confidence intervals. 
Finally, we do a perceptual emotion classification test, where humans classify the emotion of the synthesized speech.

\section{Results}

\subsection{Evaluation of Low-Resource Neutral TTS}
We first determine the possibility of training a low resource \textit{neutral} TTS system before moving on to \textit{expressive} TTS systems. 
For this, we vary the amounts of neutral data ranging from 5 hours to 30 minutes, where utterances are either randomly selected, or using the greedy algorithm to maximize syllable coverage. The greedy splits so created are incremental such that the 30 minute split is a subset of the 45 minute split and so on. We observed that using training on splits less than 1 hour, we do not obtain intelligible results from the TTS system.

\noindent \textbf{Role of Syllabic Balance:} 
We realize the importance of syllabic balance, by being able to train an intelligible TTS system with only 1 hour of greedily selected utterances, while the model trained with 1 hour of randomly selected utterances did not fare well, as indicated by metrics in Table \ref{tab:syllabic-balance}. A possible explanation for this is the fact that the random algorithm covers a mere 49\% of the unique syllables in the neutral split, in contrast to the 73\% coverage achieved by the greedy algorithm. This establishes the importance of syllabic coverage, and we use the greedy algorithm in all our further experiments.

\begin{table}[!t]
\fontsize{8pt}{10pt}\selectfont
\centering
\caption{Impact of Duration and Text Selection: 1 hour Greedy vs 1 hour Random vs. 5 hours Random in neutral TTS.}
\begingroup
\setlength{\tabcolsep}{3pt} 
\renewcommand{\arraystretch}{1} 
\begin{tabular}{@{}clrrccc@{}}
\toprule
{\begin{tabular}[c]{@{}c@{}}Duration\\ (Hours)\end{tabular}} & 
\multicolumn{1}{c}{{\begin{tabular}[c]{@{}c@{}}Text\\ Selection\end{tabular}}} & \multicolumn{1}{c}{\begin{tabular}[c]{@{}l@{}}\# \\ {syllables}\end{tabular}} & \multicolumn{1}{c}{\begin{tabular}[c]{@{}c@{}}\# {unique} \\ {syllables}\end{tabular}} & \multicolumn{1}{c}{{MCD ($\downarrow)$}} & \multicolumn{1}{c} {$F_0$ ($\downarrow$)} & \multicolumn{1}{c}{CER ($\downarrow$)} \\ \midrule
1                                                                   & Random                                                                       & 32635                                                                       & 780                                                                                & 12.54                   & 0.48                   & 0.812                  \\
1                                                                   & Greedy                                                                       & 37168                                                                       & 1163                                                                               & 10.25                   & 0.32                   & \textbf{0.074}         \\
5                                                                   & Random                                                                       & 161941                                                                      & 1343                                                                               & 10.05                   & 0.29                   & 0.085                  \\
\bottomrule
\end{tabular}
\endgroup
\label{tab:syllabic-balance}
\end{table}

\subsection{Evaluation of Expressive TTS }

\textbf{How low-resource can we go for a single emotion?}
We next investigate the minimum amount of expressive speech that is required to enable the training of expressive TTS systems. We consider a TTS system trained on 5 hours of neutral data and fine-tune on varying amounts of happy, sad, and angry speech ranging from 7.5 minutes to 1 hour. As indicated by the average MUSHRA scores for expressiveness (Exp-MUSHRA) in Table \ref{tab:angry-happy-sad-ablation}, the three aforementioned emotions start to be rated ``Fair" in  MUSHRA once we use more than or equal to 15 minutes of expressive data with the model achieving an average score. Further, by using at least 30 minutes of expressive utterances, the results fall under the ``Good" category in MUSHRA.

\begin{table}[!t]
\captionsetup{font=footnotesize}
\caption{Evaluation of TTS model built for Happy (Hap), Angry (Ang) and Sad (Sad) emotions across varying amounts of expressive data.}
\centering
\fontsize{8pt}{10pt}\selectfont 
\setlength{\tabcolsep}{3pt} 
\begin{tabular}{@{}ccccccc@{}}
\toprule
\multirow{2}{*}{\begin{tabular}[c]{@{}c@{}}Dur \\ (mins)\end{tabular}} & \multicolumn{3}{c}{CER ($\downarrow$)} & \multicolumn{3}{c}{Exp-MUSHRA($\uparrow$)} \\ \cmidrule(lr){2-4} \cmidrule(lr){5-7} 
 & Hap & Ang & Sad & Hap & Ang & Sad \\ \midrule
7.5 & 0.104 & 0.102 & 0.091 & 42.76 $\pm$ 0.91  & 38.08 $\pm$ 1.16 & 38.59 $\pm$ 1.09 \\
15 &0.087 & 0.074  & 0.075 & 53.68 $\pm$ 1.26 & 46.07 $\pm$ 0.61 & 48.30 $\pm$ 0.76 \\
30 & 0.092 & 0.089 & 0.075 & 67.96 $\pm$ 1.29  & 62.55 $\pm$ 0.67 & 61.77 $\pm$ 1.29 \\
60 &0.086 & 0.083 & 0.059 & 75.75 $\pm$ 1.01  &  62.58 $\pm$ 1.01 & 67.41 $\pm$ 1.55 \\ \bottomrule
\end{tabular}
\label{tab:angry-happy-sad-ablation}
\end{table}

\begin{table}[!t]
\fontsize{8pt}{10pt}\selectfont
\centering
\caption{Evaluation of Multi-Emotion TTS models trained on varying amounts of Neutral (N) and Expressive (E) data.}
\begingroup
\setlength{\tabcolsep}{2pt} 
\renewcommand{\arraystretch}{1} 

\begin{tabular}{@{}llllll@{}}
\toprule
\begin{tabular}[c]{@{}c@{}}\# N\\ (Hours)\end{tabular} & \begin{tabular}[c]{@{}c@{}}\# E\\ (Hours)\end{tabular} & \multicolumn{1}{c}{MCD $(\downarrow)$}          & \multicolumn{1}{c}{$F_0 (\downarrow)$}          & \multicolumn{1}{c}{CER}    & \multicolumn{1}{c}{Exp-MUSHRA} \\ \midrule
\multicolumn{1}{c}{1}                                                            & \multicolumn{1}{c}{0.5}                                                             & 13.40 $\pm$ 1.39 & 0.32 $\pm$ 0.09 & 0.103 & 57.30 $\pm$ 1.12   \\
                                                             & \multicolumn{1}{c}{1}                                                               & 13.43 $\pm$ 1.35 & 0.33 $\pm$ 0.10 & 0.097 & 59.64 $\pm$ 0.99   \\
\multicolumn{1}{c}{5}                                                            & \multicolumn{1}{c}{0.5}                                                             & 13.52 $\pm$ 1.44 & 0.32 $\pm$ 0.09 & 0.095 & 61.30 $\pm$ 1.18   \\
                                                             & \multicolumn{1}{c}{1}                                                               & 13.42 $\pm$ 1.44 & 0.32 $\pm$ 0.09 & 0.095 & 61.53 $\pm$ 1.17   \\
                                                             & \multicolumn{1}{c}{2}                                                               & 13.31 $\pm$ 1.44 & 0.33 $\pm$ 0.09 & 0.096 & 65.50 $\pm$ 1.20   \\
                                                             & \multicolumn{1}{c}{3}                                                               & 13.48 $\pm$ 1.41 & 0.33 $\pm$ 0.10 & 0.092 & 65.44 $\pm$ 1.31   \\
\multicolumn{1}{c}{10}                                                           & \multicolumn{1}{c}{0.5}                                                             & 13.30 $\pm$ 1.45 & 0.32 $\pm$ 0.09 & 0.095 & 66.06 $\pm$ 1.11   \\
                                                             & \multicolumn{1}{c}{1}                                                               & \textbf{13.19 $\pm$ 1.46} & 0.33 $\pm$ 0.10 & 0.090 & 63.54 $\pm$ 1.05   \\
                                                             & \multicolumn{1}{c}{2}                                                               & 13.45 $\pm$ 1.49 & 0.33 $\pm$ 0.09 & 0.095 & 66.41 $\pm$ 1.11   \\
                                                             & \multicolumn{1}{c}{3}                                                               & 13.34 $\pm$ 1.47 & 0.33 $\pm$ 0.09 & \textbf{0.088} & \textbf{68.66 $\pm$ 1.16}   \\ \bottomrule
\end{tabular}
\endgroup
\label{tab:main-ablation}
\end{table}


\noindent \textbf{How low-resource can we go across multiple emotions?} In the low-resource setting, we established that at least 1 hour of neutral data and 30 minutes of expressive data suffice to realize Fair TTS systems. We now focus on the high-resource setting where striking a balance between the amount of neutral and expressive data seems important. Therefore, we study the combinations of pre-training a TTS system on 1 hour, 5 hours, and 10 hours of neutral utterances and then fine-tuning the model on 30 minutes, 1 hour, 2 hours, and 3 hours of expressive speech from each emotion, respectively. We train models conditioned on all emotions (Multi) and report our findings in Table \ref{tab:main-ablation}.

\noindent \textbf{Role of Neutral Data in Expressive TTS:} In Table \ref{tab:main-ablation}, we observe the MUSHRA scores for expressiveness for different neutral TTS systems fine-tuned on increasing amounts of expressive speech. We see a trend of models pre-trained on higher neutral data achieving better expressivity even with smaller amounts of expressive data. For example, the performance of the model trained on 10 hours of neutral data and just 30 minutes of expressive data per emotion is better than the model trained on 5 hours of neutral data and 30 minutes of expressive data per emotion. Lastly, for a given neutral split, increasing the expressive data typically improves the model's expressiveness.

\begin{table}[!t]
\fontsize{8pt}{10pt}\selectfont
\centering
\caption{Subjective evaluation of Single vs. Multi-Emotion TTS models trained on varying amounts of Neutral (N) and Expressive (E) data.}
\begingroup
\setlength{\tabcolsep}{2pt} 
\renewcommand{\arraystretch}{1} 
\begin{tabular}{@{}llcccc@{}}
\toprule
\multicolumn{1}{c}{\# N} & \multicolumn{1}{c}{\begin{tabular}[c]{@{}c@{}}\# E\end{tabular}} & \multicolumn{2}{c}{CER  $(\downarrow)$}                                 & \multicolumn{2}{c}{Exp-MUSHRA $(\uparrow)$}                       \\ \cmidrule(l){3-4} \cmidrule(l){5-6}
\multicolumn{1}{l}{(Hours)} & \multicolumn{1}{l}{(Hours)} & Single                     & Multi                      & Single                           & Multi                 \\ \midrule
\multicolumn{1}{c}{1}                                                            & \multicolumn{1}{c}{0.5}                                                                                 & 0.108                     & 0.103                     & 58.64 $\pm$ 1.22                     & 57.30 $\pm$ 1.12          \\
                                                                                 & \multicolumn{1}{c}{1}                                                                                   & 0.097                     & 0.097                     & 57.75 $\pm$ 1.37                     & 59.64 $\pm$ 0.99          \\
\multicolumn{1}{c}{5}                                                            & \multicolumn{1}{c}{1}                                                                                   & 0.093                     & 0.095                     & 60.25 $\pm$ 1.29                     & 61.53 $\pm$ 1.17          \\
                                                                                 & \multicolumn{1}{c}{3}                                                                                   & \multicolumn{1}{l}{0.096} & \multicolumn{1}{l}{0.092} & \multicolumn{1}{l}{62.48 $\pm$ 1.16} & 65.44 $\pm$ 1.31          \\
\multicolumn{1}{c}{10}                                                           & \multicolumn{1}{c}{3}                                                                                   & \multicolumn{1}{l}{0.100} & \multicolumn{1}{l}{\textbf{0.088}} & \multicolumn{1}{l}{62.51 $\pm$ 0.96} & \textbf{68.66 $\pm$ 1.16} \\ \bottomrule
\end{tabular}
\endgroup
\label{tab:single-vs-multi}
\end{table}

\noindent \textbf{Comparison of Single and Multi-Emotion Models:} We also train individual models for each emotion (Single) apart from models conditioned on all emotions (Multi) and report our findings in Table \ref{tab:single-vs-multi}.  In the extreme lowest-resource setting (1 hour neutral, 0.5 hours expressive), single-emotion models marginally outperform multi-emotion ones. This is possibly due to the fact that emotional conditioning on less data creates more confusion. Our claim is also validated by the confusion matrices provided in Figure \ref{fig:cm_lr_hr}. In all other settings, we notice that multi-emotion models are better than single-emotion models. 


\subsection{Performance across Emotions} \label{sec:performance-across-emotions}

\begin{table}[!t]
\fontsize{8pt}{10pt}\selectfont 
\centering
\caption{Performance of Low-Resource (LR) and High-Resource (HR) setups across emotions.}
\begingroup
\setlength{\tabcolsep}{3pt} 
\renewcommand{\arraystretch}{1} 
\begin{tabular}{@{}lrrll@{}}
\toprule
Emotion  & \multicolumn{2}{c}{CER $(\downarrow)$} & \multicolumn{2}{c}{Exp-MUSHRA $(\uparrow)$} \\ \cmidrule(l){2-3} \cmidrule(l){4-5}
         & \multicolumn{1}{c}{LR}  & \multicolumn{1}{c}{HR} & \multicolumn{1}{c}{LR}                    & \multicolumn{1}{c}{HR}                  \\ \midrule
Happy    & 0.080                  & 0.071                 & 49.81 $\pm$ 1.35          & 61.52 $\pm$ 1.14          \\
Sad      & \textbf{0.076}         & \textbf{0.068}        & \textbf{66.79 $\pm$ 1.08} & \textbf{73.29 $\pm$ 0.73} \\
Angry    & 0.094                  & 0.081                 & 56.25 $\pm$ 1.30          & 67.55 $\pm$ 1.19          \\
Fear     & 0.122                  & 0.116                 & 53.96 $\pm$ 1.01          & 63.88 $\pm$ 1.12          \\
Disgust  & 0.115                  & 0.100                 & 60.30 $\pm$ 1.16          & 65.26 $\pm$ 1.08          \\
Surprise & 0.132                  & 0.117                 & 56.70 $\pm$ 1.19          & 61.15 $\pm$ 1.06          \\ \bottomrule
\end{tabular}
\endgroup
\label{tab:expressiveness-and-intelligibility}
\end{table}

\begin{figure}[!t]
    \centering
    \includegraphics[width=\columnwidth]{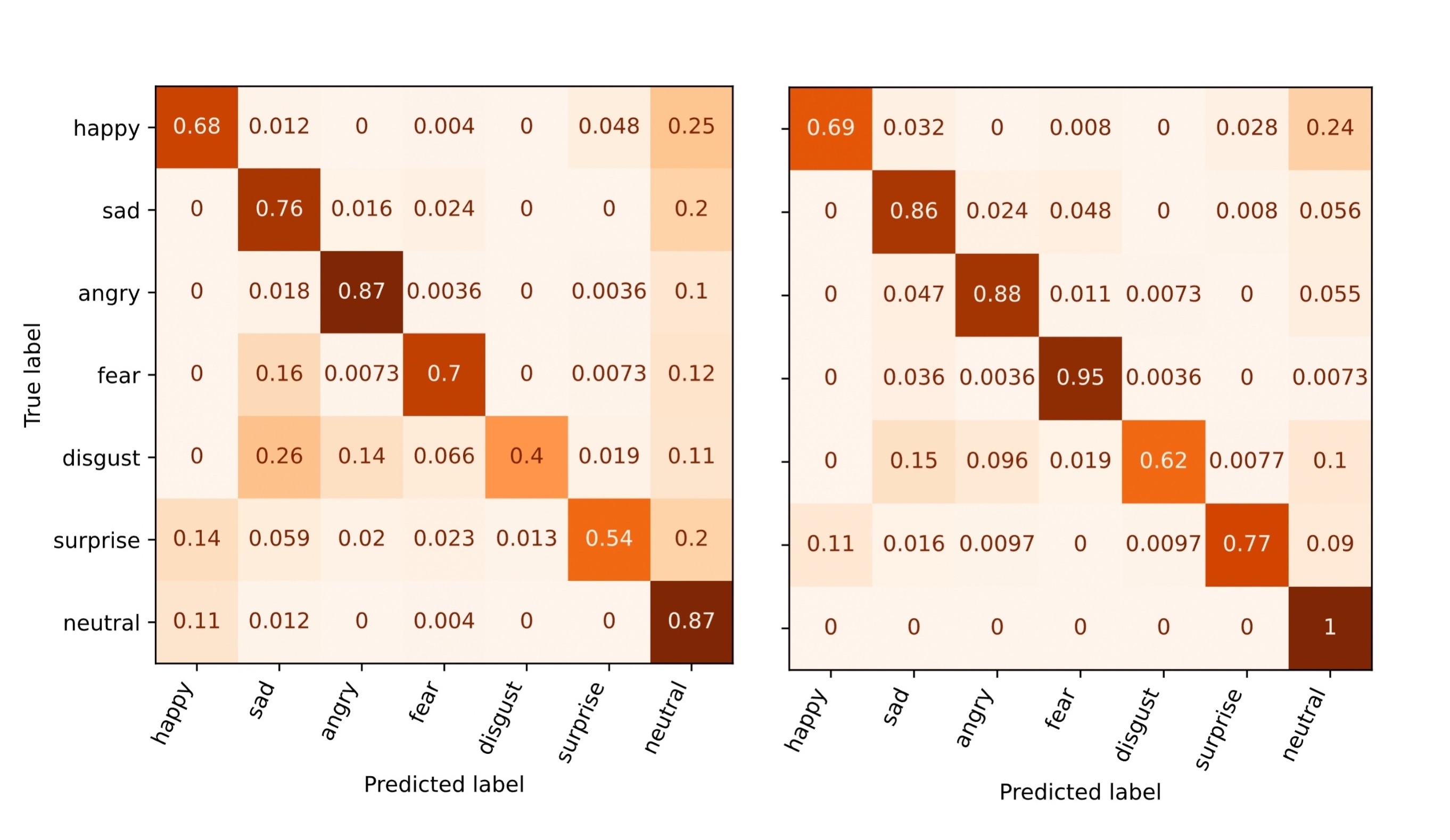}
    \caption{Confusion plots for LR (left) and HR (right) model.}
    \label{fig:cm_lr_hr}
\end{figure}

\noindent In Table \ref{tab:expressiveness-and-intelligibility}, we compare the CERs and expressiveness of our Multi-emotion models trained in (i) a low-resource setup (LR) using 1 hour of neutral and 30 minutes of expressive data and (ii) a higher-resource setup (HR) using 5 hours of neutral and 3 hours of expressive data. 
\\
\textbf{Expressiveness and Intelligibility of Our Models:} In both HR and LR setups, we notice that sad and angry emotions are simple to generate, reflected by lower CERs and higher expressiveness. With higher CERs and lower expressiveness scores, fear and surprise seem to be more difficult emotions to synthesize. This is perhaps due to the strained voice, irregular pauses, and stuttering observed while enacting these emotions. We investigate further by performing human emotion classification tests.
\\
\textbf{Perceptual Emotion Classification Results:}
The confusion matrices of emotion classification performed in human listening tests are presented in Figure \ref{fig:cm_lr_hr} for both LR and HR setups. We see that sad, anger, fear, and neutral are simpler for the model to synthesize. In the LR setting, the model struggles to produce a tone of disgust or surprise, and this trend continues in the HR setup. The listening tests provide valuable insight into understanding the challenges of building \textit{happy} TTS systems. Clearly, the emotion is often confused with neutral, and this could possibly be because the dataset naturally contains many mildly intense happy samples.

\subsection{Expressive Speech Synthesis in other Indian languages}
\label{subsec:Hindi}


We find that our results generalize well across different speaker genders and language families, when using syllabically balanced data. We train expressive male TTS for Assamese and Bengali and find that we can build Fair systems in the low-resource settings and Good systems in the high-resource settings, as indicated by Exp-MUSHRA scores in Table \ref{tab:expressiveness-and-intelligibility-for-asm-ben-him}. Next, we also re-validate our findings for building expressive TTS systems in an extremely low-resource settings. We choose Hindi, the most spoken language in India, and record one hour of neutral data and 30 minutes of expressive data with a female voice artist. Using this data we were able to 
build a Fair expressive TTS as shown by expressiveness scores in Table \ref{tab:expressiveness-and-intelligibility-for-asm-ben-him}.

\begin{table}[!t]
\fontsize{8pt}{10pt}\selectfont 
\centering
\caption{Subjective Evaluation of expressive TTS for Assamese (as), Bengali (bn), and Hindi (hi).}
\begingroup
\setlength{\tabcolsep}{3pt} 
\renewcommand{\arraystretch}{1} 
\begin{tabular}{@{}lcclc@{}}
\toprule
Language & \begin{tabular}[l]{@{}c@{}}\# N\\ (Hours)\end{tabular} & \begin{tabular}[l]{@{}c@{}}\# E\\ (Hours)\end{tabular} & Exp-MUSHRA ($\uparrow$) \\ \midrule
Hindi             & 1                                                               & 0.5                                                             & 49.69 $\pm$ 1.23       \\
Assamese          & 1                                                               & 0.5                                                             & 51.85 $\pm$ 0.63       \\
                  & 10                                                              & 1                                             & 60.04 $\pm$ 0.64       \\
Bengali           & 1                                                               & 0.5                                                             & 49.58 $\pm$ 0.60       \\
                  & 10                                                              & 1                                             & 73.49 $\pm$ 0.83       \\ \bottomrule
\end{tabular}
\endgroup
\label{tab:expressiveness-and-intelligibility-for-asm-ben-him}
\end{table}

\section{Related Work}

\textbf{Low-Resource TTS:}
Phonetic balance is an important aspect of designing TTS corpora  \cite{kominek2004cmu, Srivastava2020IndicSpeech}. 
Apart from phonetic balance, other effective data-balancing strategies \cite{Yang2020Towards, Seki2023}  can help mitigate imbalance when training with both high and low-resource languages. Another interesting theme is the adaptation of resource-rich models to low-resource setups as shown in multi-speaker \cite{neekhara2021adapting} and multilingual \cite{prakash2020generic, He2021Multilingual, elneima2022adversarial, Lux2022} contexts. 
E-TTS \cite{gupta2023tts} introduces a novel data augmentation technique leveraging minimal prosodically-rich data to enhance intelligibility and expressiveness in low-resource TTS systems for Hindi.
In this work, we focus on the low-resource problem  of building expressive TTS models by adapting a neutral TTS model using a smaller amount of multi-emotion data.\\
\textbf{Expressive TTS:}
Several existing works build expressive neural speech synthesizers, by conditioning the model on discrete emotion labels
\cite{lorenzo-trueba_investigating_2018, lee_emotional_2017} or on reference speech \cite{wang_style_2018}. Many recent works \cite{li2021towards, guo2022emodiff, lei2022msemotts} effectively build expressive TTS systems but in high-resource settings. In the context of low-resource expressive TTS systems, few works \cite{Huybrechts2021Low-Resource, terashima2022cross}, show that voice conversion can aid cross-speaker transfer.
However, such setups have at least one speaker containing fair amounts of expressive data, as opposed to our low-resource setting. \\
\textbf{TTS Datasets:} The few existing publicly available emotion-labelled speech corpora \cite{Busso2008IEMOCAP, lemoine2020Att, Zhou2021Seen, Cui2021EMOVIE} do not cover Indian languages. Datasets such as IndicTTS \cite{baby2016resources} and IndicSpeech \cite{Srivastava2020IndicSpeech} contain only neutral voices for some Indian languages. 
\section{Conclusion}
In this work, we release \textit{Rasa}, a dataset for building expressive TTS systems in Assamese, Bengali, and Tamil containing 10 hours of neutral speech and 1-3 hours of expressive speech for the 6 basic Ekman emotions. Using this data we show that we can build Fair expressive TTS systems using only 1 hour of neutral data and 30 minutes of expressive data per emotion. We further establish that using more neutral data reduces the requirement of expressive data per emotion. Our ablation studies provided insights about (i) the minimum amount of expressive data needed per emotion (ii) the difficulty of synthesizing certain emotions, and (iii) the better expressivity of multi-emotion systems over single-emotion systems. We believe that our findings will provide a blueprint for collecting datasets for low-resource languages, appropriately  balancing neutral and expressive data.

\noindent \textbf{Acknowledgements} We  thank Digital India Bhashini, the Ministry of Electronics and Information Technology of the Government of India, Centre for Development of Advanced Computing Pune, EkStep Foundation and Nilekani Philanthropies for their generous grants and support. We also thank the entire team at AI4Bharat, especially Mohanarangan A, Krishnan K S, Janki Nawale, and Ishvinder Sethi.
\bibliographystyle{IEEEtran}
\bibliography{refs_remove_url_clean_long_names}

\begin{thebibliography}{10}
\providecommand{\url}[1]{#1}
\csname url@samestyle\endcsname
\providecommand{\newblock}{\relax}
\providecommand{\bibinfo}[2]{#2}
\providecommand{\BIBentrySTDinterwordspacing}{\spaceskip=0pt\relax}
\providecommand{\BIBentryALTinterwordstretchfactor}{4}
\providecommand{\BIBentryALTinterwordspacing}{\spaceskip=\fontdimen2\font plus
\BIBentryALTinterwordstretchfactor\fontdimen3\font minus \fontdimen4\font\relax}
\providecommand{\BIBforeignlanguage}[2]{{%
\expandafter\ifx\csname l@#1\endcsname\relax
\typeout{** WARNING: IEEEtran.bst: No hyphenation pattern has been}%
\typeout{** loaded for the language `#1'. Using the pattern for}%
\typeout{** the default language instead.}%
\else
\language=\csname l@#1\endcsname
\fi
#2}}
\providecommand{\BIBdecl}{\relax}
\BIBdecl

\bibitem{Busso2008IEMOCAP}
C.~Busso, M.~Bulut, C.~Lee, A.~Kazemzadeh, E.~Mower, S.~Kim, J.~N. Chang, S.~Lee, and S.~S. Narayanan, ``{IEMOCAP:} interactive emotional dyadic motion capture database,'' \emph{Lang. Resour. Evaluation}, vol.~42, no.~4, pp. 335--359, 2008.

\bibitem{lemoine2020Att}
C.~Le~Moine and N.~Obin, ``{Att-HACK: An Expressive Speech Database with Social Attitudes},'' in \emph{{Speech Prosody}}, Tokyo, Japan, May 2020.

\bibitem{Zhou2021Seen}
K.~Zhou, B.~Sisman, R.~Liu, and H.~Li, ``Seen and unseen emotional style transfer for voice conversion with {A} new emotional speech dataset,'' in \emph{{IEEE} {ICASSP} 2021, Toronto, ON, Canada, June 6-11, 2021}.\hskip 1em plus 0.5em minus 0.4em\relax {IEEE}, 2021, pp. 920--924.

\bibitem{Cui2021EMOVIE}
C.~Cui, Y.~Ren, J.~Liu, F.~Chen, R.~Huang, M.~Lei, and Z.~Zhao, ``{EMOVIE:} {A} mandarin emotion speech dataset with a simple emotional text-to-speech model,'' in \emph{INTERSPEECH 2021}.\hskip 1em plus 0.5em minus 0.4em\relax {ISCA}, 2021, pp. 2766--2770.

\bibitem{nguyen23}
T.~A. Nguyen, W.-N. Hsu, A.~D'Avirro, B.~Shi, I.~Gat, M.~Fazel-Zarani, T.~Remez, J.~Copet, G.~Synnaeve, M.~Hassid, F.~Kreuk, Y.~Adi, and E.~Dupoux, ``{Expresso: A Benchmark and Analysis of Discrete Expressive Speech Resynthesis},'' in \emph{Proc. INTERSPEECH 2023}, 2023, pp. 4823--4827.

\bibitem{Cowen2017}
A.~S. Cowen and D.~Keltner, ``Self-report captures 27 distinct categories of emotion bridged by continuous gradients,'' \emph{Proceedings of the National Academy of Sciences}, vol. 114, no.~38, pp. E7900--E7909, 2017.

\bibitem{ekman1992argument}
P.~Ekman, ``An argument for basic emotions,'' \emph{Cognition \& emotion}, vol.~6, no. 3-4, pp. 169--200, 1992.

\bibitem{baby2016resources}
A.~Baby, A.~L. Thomas, N.~Nishanthi, T.~Consortium \emph{et~al.}, ``Resources for indian languages,'' in \emph{Proceedings of Text, Speech and Dialogue}, 2016.

\bibitem{Black2019CMU}
A.~W. Black, ``{CMU} wilderness multilingual speech dataset,'' in \emph{{IEEE} {ICASSP} 2019}.\hskip 1em plus 0.5em minus 0.4em\relax {IEEE}, 2019, pp. 5971--5975.

\bibitem{Srivastava2020IndicSpeech}
N.~Srivastava, R.~Mukhopadhyay, K.~R. Prajwal, and C.~V. Jawahar, ``Indicspeech: Text-to-speech corpus for indian languages,'' in \emph{{LREC} 2020}.\hskip 1em plus 0.5em minus 0.4em\relax European Language Resources Association, 2020, pp. 6417--6422.

\bibitem{lancucki2021fastpitch}
A.~{\L}a{\'n}cucki, ``Fastpitch: Parallel text-to-speech with pitch prediction,'' in \emph{ICASSP 2021}.\hskip 1em plus 0.5em minus 0.4em\relax IEEE, 2021, pp. 6588--6592.

\bibitem{badlani2022one}
R.~Badlani, A.~{\L}a{\'n}cucki, K.~J. Shih, R.~Valle, W.~Ping, and B.~Catanzaro, ``One tts alignment to rule them all,'' in \emph{ICASSP 2022}.\hskip 1em plus 0.5em minus 0.4em\relax IEEE, 2022, pp. 6092--6096.

\bibitem{kong2020hifi}
J.~Kong, J.~Kim, and J.~Bae, ``Hifi-gan: Generative adversarial networks for efficient and high fidelity speech synthesis,'' \emph{Advances in Neural Information Processing Systems}, vol.~33, pp. 17\,022--17\,033, 2020.

\bibitem{khan2024indicllmsuite}
M.~S. U. R.~K. et. al., ``Indicllmsuite: A blueprint for creating pre-training and fine-tuning datasets for indian languages,'' 2024.

\bibitem{gala2023}
J.~Gala~et. al., ``Indictrans2: Towards high-quality and accessible machine translation models for all 22 scheduled indian languages,'' \emph{TMLR}, 2023.

\bibitem{kunchukuttan2020indicnlp}
A.~Kunchukuttan, ``{The IndicNLP Library},'' \url{https://github.com/anoopkunchukuttan/indic_nlp_library/blob/master/docs/indicnlp.pdf}, 2020.

\bibitem{Kumar2022Towards}
G.~K.~K. et. al., ``Towards building text-to-speech systems for the next billion users,'' in \emph{{IEEE} {ICASSP} 2023}.\hskip 1em plus 0.5em minus 0.4em\relax {IEEE}, 2023, pp. 1--5.

\bibitem{vaswani2017attention}
A.~Vaswani, N.~Shazeer, N.~Parmar, J.~Uszkoreit, L.~Jones, A.~N. Gomez, {\L}.~Kaiser, and I.~Polosukhin, ``Attention is all you need,'' \emph{Advances in neural information processing systems}, vol.~30, 2017.

\bibitem{santen1997methods}
J.~P.~H. van Santen and A.~L. Buchsbaum, ``Methods for optimal text selection,'' in \emph{{EUROSPEECH} 1997}.\hskip 1em plus 0.5em minus 0.4em\relax {ISCA}, 1997, pp. 553--556.

\bibitem{kubichek1993mel}
R.~Kubichek, ``Mel-cepstral distance measure for objective speech quality assessment,'' in \emph{Proceedings of IEEE pacific rim conference on communications computers and signal processing}, vol.~1.\hskip 1em plus 0.5em minus 0.4em\relax IEEE, 1993, pp. 125--128.

\bibitem{salvador2007toward}
S.~Salvador and P.~Chan, ``Toward accurate dynamic time warping in linear time and space,'' \emph{Intelligent Data Analysis}, vol.~11, no.~5, pp. 561--580, 2007.

\bibitem{series2014method}
B.~Series, ``Method for the subjective assessment of intermediate quality level of audio systems,'' \emph{ITU-R}, 2014.

\bibitem{schoeffler2018webmushra}
M.~Schoeffler, S.~Bartoschek, F.-R. St{\"o}ter, M.~Roess, S.~Westphal, B.~Edler, and J.~Herre, ``webmushra—a comprehensive framework for web-based listening tests,'' \emph{Journal of Open Research Software}, vol.~6, no.~1, p.~8, 2018.

\bibitem{kominek2004cmu}
J.~Kominek and A.~W. Black, ``The {CMU} arctic speech databases,'' in \emph{Fifth {ISCA} {ITRW} on Speech Synthesis, Pittsburgh, PA, USA, June 14-16, 2004}.\hskip 1em plus 0.5em minus 0.4em\relax {ISCA}, 2004, pp. 223--224.

\bibitem{Yang2020Towards}
J.~Yang and L.~He, ``Towards universal text-to-speech,'' in \emph{INTERSPEECH 2020}.\hskip 1em plus 0.5em minus 0.4em\relax {ISCA}, 2020, pp. 3171--3175.

\bibitem{Seki2023}
K.~Seki, S.~Takamichi, T.~Saeki, and H.~Saruwatari, ``Diversity-based core-set selection for text-to-speech with linguistic and acoustic features,'' \emph{CoRR}, vol. abs/2309.08127, 2023.

\bibitem{neekhara2021adapting}
P.~Neekhara, J.~Li, and B.~Ginsburg, ``Adapting {TTS} models for new speakers using transfer learning,'' \emph{CoRR}, vol. abs/2110.05798, 2021.

\bibitem{prakash2020generic}
A.~Prakash and H.~A. Murthy, ``Generic indic text-to-speech synthesisers with rapid adaptation in an end-to-end framework,'' in \emph{INTERSPEECH 2020}.\hskip 1em plus 0.5em minus 0.4em\relax {ISCA}, 2020, pp. 2962--2966.

\bibitem{He2021Multilingual}
M.~He, J.~Yang, L.~He, and F.~K. Soong, ``Multilingual byte2speech models for scalable low-resource speech synthesis,'' \emph{CoRR}, vol. abs/2103.03541, 2021.

\bibitem{elneima2022adversarial}
A.~Elneima and M.~Binkowski, ``Adversarial text-to-speech for low-resource languages,'' in \emph{WANLP@EMNLP 2022}.\hskip 1em plus 0.5em minus 0.4em\relax Association for Computational Linguistics, 2022, pp. 76--84.

\bibitem{Lux2022}
F.~Lux, J.~Koch, and N.~T. Vu, ``Low-resource multilingual and zero-shot multispeaker {TTS},'' in \emph{AACL, IJCNLP 2022}.\hskip 1em plus 0.5em minus 0.4em\relax Association for Computational Linguistics, Nov. 2022, pp. 741--751.

\bibitem{gupta2023tts}
I.~Gupta and H.~A. Murthy, ``E-tts: Expressive text-to-speech synthesis for hindi using data augmentation,'' in \emph{International Conference on Speech and Computer}.\hskip 1em plus 0.5em minus 0.4em\relax Springer, 2023, pp. 243--257.

\bibitem{lorenzo-trueba_investigating_2018}
J.~Lorenzo-Trueba, G.~Eje~Henter, S.~Takaki, J.~Yamagishi, Y.~Morino, and Y.~Ochiai, ``\BIBforeignlanguage{en}{Investigating different representations for modeling and controlling multiple emotions in {DNN}-based speech synthesis},'' \emph{\BIBforeignlanguage{en}{Speech Communication}}, vol.~99, pp. 135--143, May 2018.

\bibitem{lee_emotional_2017}
Y.~Lee, A.~Rabiee, and S.-Y. Lee, ``Emotional {End}-to-{End} {Neural} {Speech} {Synthesizer},'' in \emph{ML4Audio Workshop at NIPS 2017}, 2017.

\bibitem{wang_style_2018}
Y.~Wang, D.~Stanton, Y.~Zhang, R.~J. Skerry-Ryan, E.~Battenberg, J.~Shor, Y.~Xiao, F.~Ren, Y.~Jia, and R.~A. Saurous, ``Style {Tokens}: {Unsupervised} {Style} {Modeling}, {Control} and {Transfer} in {End}-to-{End} {Speech} {Synthesis},'' Mar. 2018, arXiv:1803.09017 [cs, eess].

\bibitem{li2021towards}
X.~Li, C.~Song, J.~Li, Z.~Wu, J.~Jia, and H.~Meng, ``Towards multi-scale style control for expressive speech synthesis,'' in \emph{INTERSPEECH 2021}.\hskip 1em plus 0.5em minus 0.4em\relax {ISCA}, 2021, pp. 4673--4677.

\bibitem{guo2022emodiff}
Y.~Guo, C.~Du, X.~Chen, and K.~Yu, ``Emodiff: Intensity controllable emotional text-to-speech with soft-label guidance,'' in \emph{ICASSP 2023}, 2023, pp. 1--5.

\bibitem{lei2022msemotts}
Y.~Lei, S.~Yang, X.~Wang, and L.~Xie, ``Msemotts: Multi-scale emotion transfer, prediction, and control for emotional speech synthesis,'' \emph{{IEEE} {ACM} TASLP}, vol.~30, pp. 853--864, 2022.

\bibitem{Huybrechts2021Low-Resource}
G.~Huybrechts, T.~Merritt, G.~Comini, B.~Perz, R.~Shah, and J.~Lorenzo{-}Trueba, ``Low-resource expressive text-to-speech using data augmentation,'' in \emph{{IEEE} {ICASSP} 2021}.\hskip 1em plus 0.5em minus 0.4em\relax {IEEE}, 2021, pp. 6593--6597.

\bibitem{terashima2022cross}
R.~Terashima, R.~Yamamoto, E.~Song, Y.~Shirahata, H.~Yoon, J.~Kim, and K.~Tachibana, ``Cross-speaker emotion transfer for low-resource text-to-speech using non-parallel voice conversion with pitch-shift data augmentation,'' in \emph{Interspeech 2022}.\hskip 1em plus 0.5em minus 0.4em\relax {ISCA}, 2022, pp. 3018--3022.

\end{thebibliography}

\end{document}